\documentclass[12pt]{article}

\usepackage{arxiv}
\usepackage[utf8]{inputenc} 
\usepackage[T1]{fontenc}    
\usepackage{hyperref}       
\usepackage{url}            
\usepackage{booktabs}       
\usepackage{amsfonts}       
\usepackage{nicefrac}       
\usepackage{microtype}      
\usepackage{lipsum}

\usepackage{color}

\usepackage{graphics}
\usepackage{graphicx}
\usepackage{amsmath}
\usepackage{amssymb}
\usepackage{upgreek}
\usepackage{cite}
\usepackage{refstyle}
\usepackage{caption}
\usepackage{subcaption}
\usepackage{float}
\usepackage{fullpage}
\usepackage{tikz}
\usepackage{tkz-berge}
\usepackage{listings}
\usepackage{commath}
\usepackage{setspace}

\title{Unsupervised Videographic Analysis of Rodent Behaviour.}

\author{
  Anthony Bourached\\
  Department of Computer Science\\
  University College London\\
  London \\
  \texttt{anthony.bourached.18@ucl.ac.uk} \\
   \And
  Parashkev Nachev \\
  Institue of Neurology\\
  University College London\\
  London \\
  \texttt{p.nachev@ucl.ac.uk} \\
}

\begin{document}

\newpage

\maketitle

\begin{abstract}
    Animal behaviour is complex and the amount of data in the form of video, if extracted, is copious. Manual analysis of behaviour is massively limited by two insurmountable obstacles, the complexity of the behavioural patterns and human bias. Automated visual analysis has the potential to eliminate both of these issues and also enable continuous analysis allowing a much higher bandwidth of data collection which is vital to capture complex behaviour at many different time scales. Behaviour is not confined to a finite set modules and thus we can only model it by inferring the generative distribution. In this way unpredictable, anomalous behaviour may be considered. Here we present a method of unsupervised behavioural analysis from nothing but high definition video recordings taken from a single, fixed perspective. We demonstrate that the identification of stereotyped rodent behaviour can be extracted in this way. 
\end{abstract}

\section{Introduction}

\paragraph{}
The need for automated and efficient systems for tracking full animal pose has increased with the complexity of behavioural data and analyses. One of the most prominent objectives enabled by this is behavioural analysis. 

\paragraph{}
It is not possible to know a priori what time scale is most relevant to analyse complex behaviour nor is manual analysis satisfactory to capture every aspect of this behaviour. Indeed, it is considered that animal behaviours are likely built from simple modules, and that their systematic identification is the real challenge. Wiltschko et al \cite{mousebehaviour} show that for mice these finite modules are defined on sub-second time scales and that complex behaviour is formed from a string of these stereotyped behaviours concatenated according to transition probabilities. However, automatic behavioural analysis poses many problems. First, an 'understanding' of the relevant components that generate behaviour, namely the animals themselves- and their bodyparts, is required prior to any useful analysis. Thus behavioural analysis foundationally depends on machine vision: object detection, segmentation and more specifically pose extraction. \cite{mouse_light_dark} conducts a comprehensive comparison of non-video and non-invasive ways of analysing rodent behaviour. This involves technology to measure number of wheel turns, licks of a water bottle and other metrics derived from other such objects that a rodent will interact with in its home cage environment. Although metrics like wheel turning have been widely utilised to determine circadian rhythms and sleep \cite{mouse_light_dark} they are all insufficient for analysing more garnular forms of behaviour.

\paragraph{}
A baseline approach to pose extraction may involve attaching sensors to a subject and using these to extract the positions of the locations of interest as they evolve over time. This however clearly can not be done without significantly influencing the normal behaviour of the test subject. Furthermore, it rules out many applications of such technology even if complex behaviour can be derived, and categorised in this way. In \cite{mouse_automated} Bains et al analyse activity in mice by using radio-frequency identification implants. Although administered humanely and full recovery was allowed before analysis was conducted this procedure can not be considered non invasive and any such procedure could not be used inferencially. Thus much of modern day machine vision has turned its attention to pose estimation where deep learning techniques, almost always, involving in some fashion the use of deep convolutional neural networks are used to learn categorisation of objects in bounding boxes (object detection), and pixels in an object (object segmentation). In this way, not just animals but body parts may be located in an image.

\paragraph{}
In recent years this has been particularly successful in human pose estimation \cite{human_pose_estimation}, \cite{deeppose}. Particularly, openpose \cite{openpose} takes this a step further. Using a bottom-up approach\footnote{Here we use bottom-up in the same capacity as Cao et al in meaning first detecting bodyparts then inferring a pose rather than top-down where the whole body would first be detected.} allows multiple highly similar features to exist in the same image and be identified independently without suffering from early commitment associated with top-down approaches. This is especially important when there are multiple people in close proximity since it is the case where person detectors are most likely to fail. Densepose, Güler et al \cite{densepose}, presents a full-blown supervised method of learning a correspondence between 2D RGB images and detailed, accurate parametric surface model of the human body in 3D. Unfortunately some of these advanced techniques have not yet been implemented open source for other animals. In general, the task of object detection and segmentation across a wide variety of objects can be achieved by transfer learning with a minimal amount of labelled data\footnote{We train here a novel rodent detection and segmentation algorithm using transfer learning and less than 100 labelled images.} however pose estimation is far more difficult to achieve even with excellent data and is a major obstacle to achieving automatic behavioural analysis.

\subsection{Introduction to Methodology}

\paragraph{Objective}
Here we seek to demonstrate such finite stereotyped behaviour similar to that shown by \cite{mousebehaviour} in an unsupervised way for rodents in their home cage environment. Labelling shall only be necessary for the purpose of body and body-part detection. Both of these will be achieved using transfer learning as discussed below. All behavioural analysis thereafter, particularly for short timescales (one or two seconds), shall be unsupervised and any stereotyped behaviour found will be visualised.

\subsubsection{Pose Estimation}

\paragraph{Object Detection}
We shall train object detection using the MaskRCNN \cite{maskrcnn} architecture to perform transfer learning with the COCO (Common Objects in Context) weights \cite{matterport_maskrcnn_2017}. A small labelled subset of rodent images from our dataset will be used for training and validation. 

\paragraph{Spotlight Videos}
From here this object detection model will be used to extract 'spotlight' videos wherein the objective is to have exactly one rodent in view for the duration of each video. This process will enable significantly better pose estimation. This will be further motivated and discussed in section \ref{sec:spotlight}. Next we shall use a pose estimation toolkit, DeepLabCut, developed by Mathis et al \cite{deeplabcut} to learn a set of key bodyparts considered relevant for behavioural analysis. This is achieved by labelling these bodyparts in a small set of frames extracted from videos in our dataset using k-means clustering and using transfer learning from the default deeplabcut weights. Deeplabcut has a comprehensive guide to achieve this, we followed \cite{deeplabcut_user}. Finally, using a threshold confidence and custom designed metrics we shall extract the highest quality spotlight videos (defined at a high level as having minimum occlusions and maximum joint bodypart detection confidence across the videos). This concludes the pose extraction process, resulting in time series data which requires cleaning and interpolation.

\subsubsection{Behavioural Analysis from Time Series}

\paragraph{Interpolation}
Differential analysis, herein referred to as differtial smoothing, shall be used to remove spurious points from the time series that will then be replaced by more probable values during interpolation. We will experiment with linear interpolation and cubic splines with and without differential smoothing. 

\paragraph{Manifold Projections}
We shall then extract windows of a fixed number of frames, $\omega$, with the spatial coordinates of the bodyparts of interest included for each frame. These windows will be selected from each video with a temporal stride of $s=1$. Using Uniform Manifold Approximation and Projection \cite{umap} we learn a projection of these windows onto two dimensional space such that the resulting datapoints each represent a complex nonlinear relationship between the joint bodypart of interest postions and time. This will be compared to a baseline of linear dimensionality reduction to demonstrate the significance of the capturing of complex, non-linear behaviour. Space in this projection thus characterises difference in behaviour. It follows that clusters of points on this projection represents the stereotyped behaviour of interest which we shall visually characterise by overlaying the original edge enhanced and thresholded frames corresponding to several spatial regions.

\subsection{Data}

\paragraph{}
The raw data for this project was provided in the form of several hundred of hours of rodents in their home cage environment. The age, gender and colour of the rodents varied throughout the dataset, however the cage size, set-up and camera position remained constant for all videos. All data was provided courtesy of the MRC Harwell Institute, using the Home Cage Analysis System developed by Actual Analytics, Edinburgh.

\section{Pose Estimation}
\label{sec:spotlight}

Since the objective is to extract the positions of the subjects' bodyparts in each frame it would be ideal to use deeplabcut on the raw video image. However, this presented multiple insurmountable obstacles as outlined below.

\subsection{Problem with using DeepLabCut on raw data}
\label{sec:problems_dlc}

Parametising using DeepLabCut on raw video resulted in poor parametisation for the following primary reasons:

\begin{itemize}
    \item The rodent cage floor is made of pencil-shaving like bedding and other movable toys which often occludes multiple body parts making most of the video unuseful, since body part positions are indeterminable. This issue could be improved and better data efficiency achieved by more careful video recording methodology.
    \item At the time of simulation the current deeplabcut does not have a body model and so using it alone makes it impossible to tell which body parts belong to which rodents which wouldn’t even be solved well by having a body model as they are often in very close proximity.
\end{itemize}

\begin{figure}[H]
	\centering
	\begin{subfigure}[h]{0.45\textwidth}
		\centering
		\includegraphics[width=\textwidth]{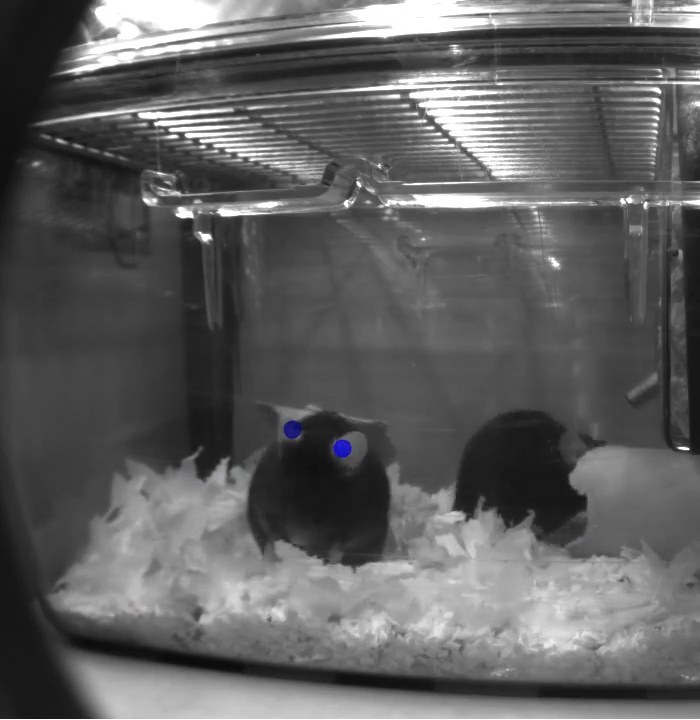}
		\caption{Two ears labelled correctly.}
		\label{subfig:two_ears}
	\end{subfigure}
	\hfill
	\begin{subfigure}[h]{0.45\textwidth}
		\centering
		\includegraphics[width=\textwidth]{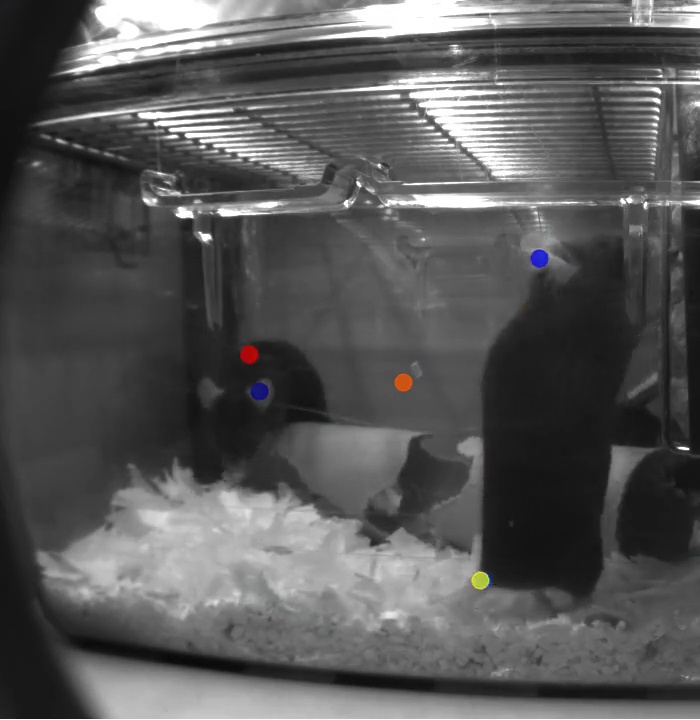}
		\caption{Two ears labelled correctly but on different rodents.}
		\label{subfig:ears_diff}
	\end{subfigure} 
	\caption{Issue with using deeplabcut on raw video data- with multiple rodents in view at once. Hence motivating the necessity of creating spotlight videos.}
	\label{fig:deeplabcut_issues}
\end{figure}

\subsection{Contiguous frame selection for spotlight videos}

\paragraph{}
It was found that both these issues could be mitigated by extracting zoomed in videos of the rodents where there is only one rodent that is visible for the duration of the video. This was achieved by labelling a small dataset of rodents from a mix of images from our videos and from online sources and using these for transfer learning with MaskRCNN \cite{maskrcnn} with the learned COCO. For this we followed \cite{matterport_maskrcnn_2017}.

\begin{figure}[H]
	\centering
	\begin{subfigure}[h]{0.45\textwidth}
		\centering
		\includegraphics[width=\textwidth]{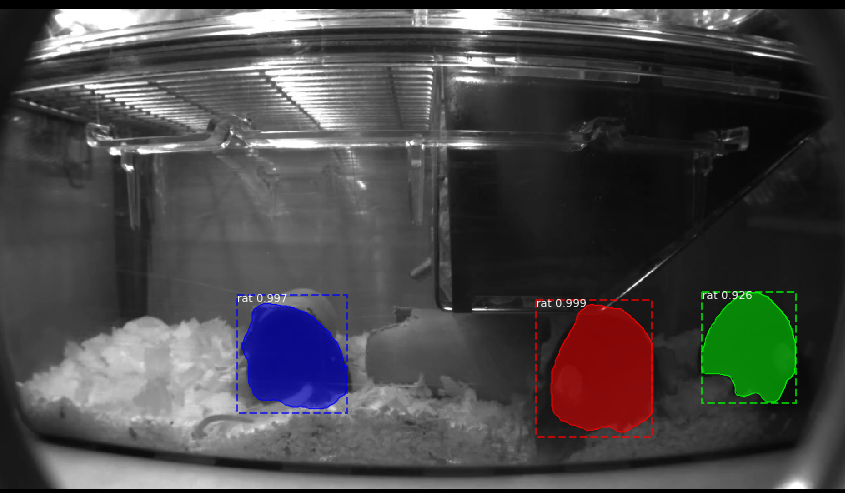}
		\caption{Object detection, no collision. This frame may contribute to potentially three spotlight videos.}
		\label{subfig:detection}
	\end{subfigure}
	\hfill
	\begin{subfigure}[h]{0.45\textwidth}
		\centering
		\includegraphics[width=\textwidth]{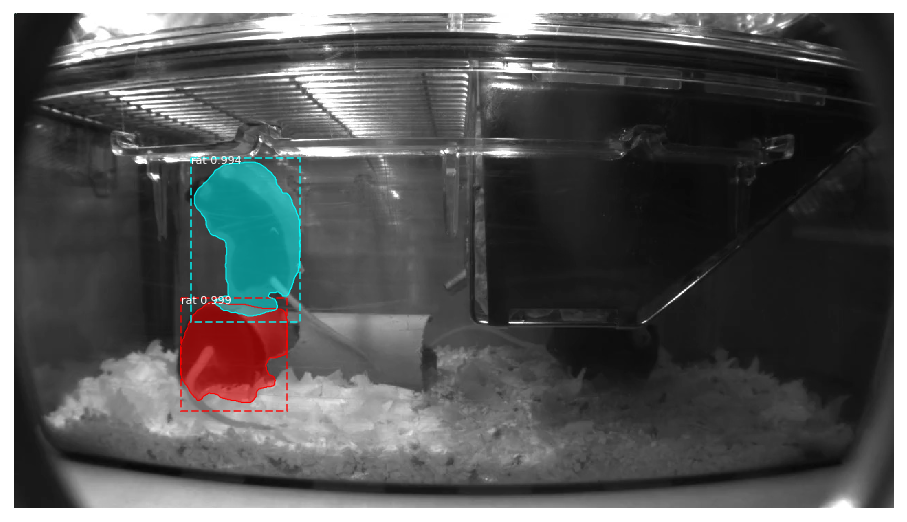}
		\caption{Bounding box collision. Due to this collision neither of these bounding boxes may contribute to a spotlight video.}
		\label{subfig:collision}
	\end{subfigure} 
	\caption{Object detection with transfer learned rodent weights. Each non-colliding bounding box will contribute to a spotlight video if the boxes' centres are within a threshold $\epsilon$ from the centre of a bounding box in the antecedent frame that is not already part of a contiguous set of spotlights.}
	\label{fig:object_detections}
\end{figure}

\paragraph{}
Bounding boxes are then extracted from sets of contiguous frames that are given high probability of containing the same rodent while also removing collisions. In this way high quality, spotlight, videos could be formed that minimises occlusions and collisions.

\paragraph{}
Above we have specified a mechanism through which we intend to create spotlight videos. How wide should the spotlight around the subject be? By what euclidean distance may the centre of a bounding box move from one frame to the next? What threshold confidence in the bounding boxes should be used? These are all hyperparameters of the spotlight video algorithm that needed to be determined and are considered below.

\subsubsection{Threshold confidence = 0.75}

\paragraph{}
The threshold confidence for determining if a rodent was in a bounding box was a difficult decison to make. If this threshold was too low then the probability of selecting a video containing no rodents is increased. However, if it is too high the number of videos selected becomes very low- especially since at least 50 contiguous frames where the same bounding box\footnote{By same bounding box we mean the bounding box belonging to the same rat.} is detected was required to make 2 seconds of video. The number of collisions detected also decreases as the confidence threshold increases. 

\paragraph{}
Values examined for this hyperparameter were $0.95$, $0.75$ and $0.5$. $0.75$ was accepted as it produced almost as many videos as 0.5 while rarely making videos of non-rodents.

\subsubsection{Bounding box grace}

$\delta = 50$

\paragraph{}
$\delta$ is the number of pixels added to each edge of the bounding box when extracting the cropped videos. Without this outstanding body parts such as snout, paws and ears often exited the edge of the bounding box and thus could not be tracked in the cropped video. $\delta = 50$ was found to be sufficiently large to make this a very rare occurrence while not being too large to drastically increase the number of times other rodents invaded the cropped video.

\subsubsection{Threshold distance}

$\epsilon = 50.0$

\paragraph{}
To determine if a bounding box is tracking the same object from frame to frame we define some small distance $\epsilon$. If the distance between the centre of a bounding box in frame j is less than some small distance $\epsilon$ from the centre of a bounding box in the preceding frame i then it is labelled as the same bounding box. This value of $\epsilon$ is measured in euclidean distance. Small values of $\epsilon$ proved unreliable for three reasons: 

\begin{itemize}
    \item Some of the interesting movements are fast and so the bounding box can move a considerable amount from frame to frame. 
    \item Posture changes as well as changing position and also changes in the position of the centre of the box since the centre moves as the shape and size of the box moves. 
    \item The object detection is done on each frame independently and so there is significant variance in the shape and size of the bounding box especially when fast motion is involved which compounds the significance of the first two issues.
\end{itemize}

A value of $\epsilon = 50.0$ was sufficiently large to capture almost all fast motions and was still sufficiently small that non-colliding bounding boxes were rarely within a distance of $2 \epsilon$ from each other.

\begin{figure}[H]
	\centering
	\begin{subfigure}[h]{0.45\textwidth}
		\centering
		\includegraphics[width=\textwidth]{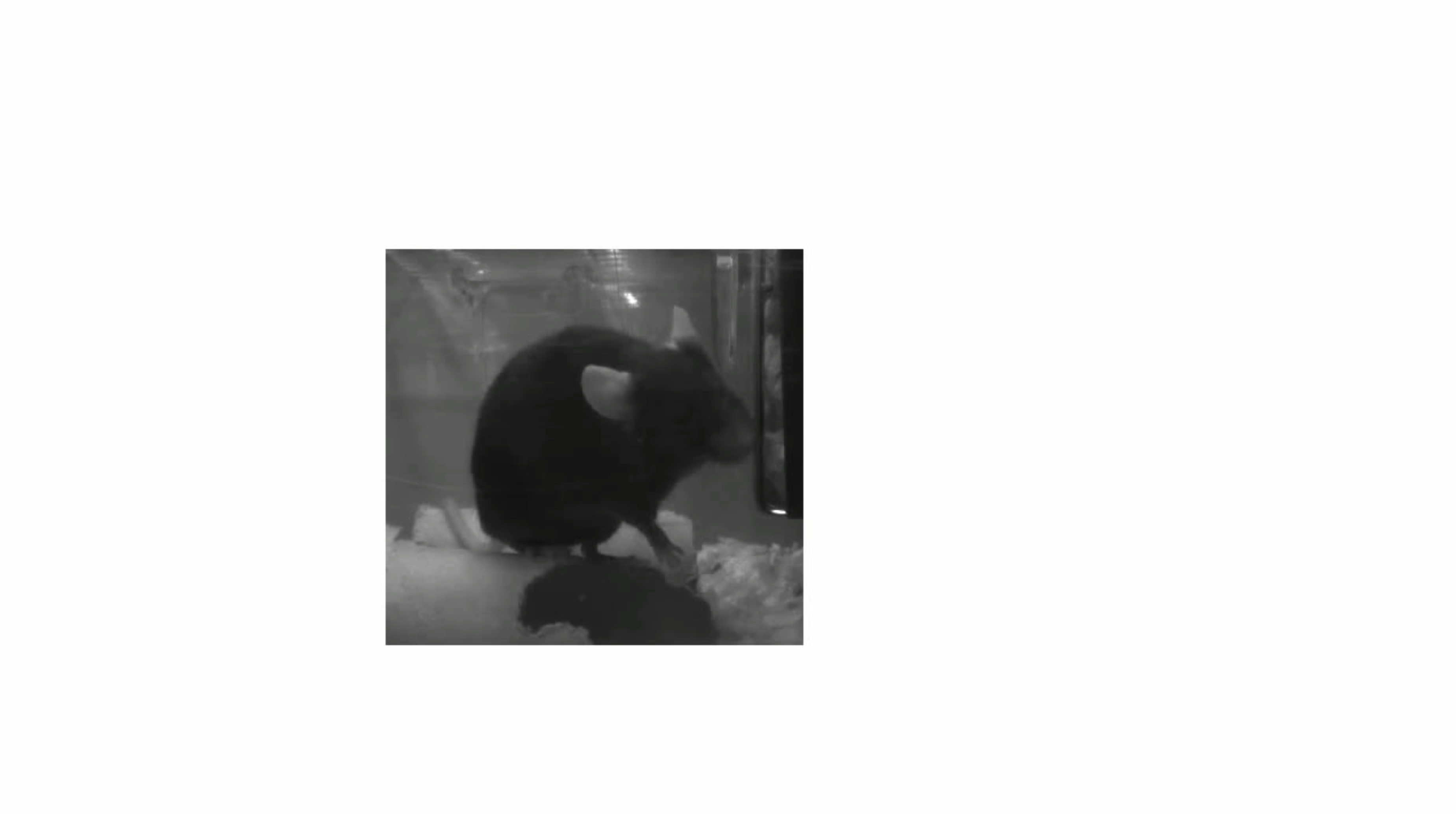}
		\caption{Example 1.}
		\label{subfig:vid_exm1}
	\end{subfigure}
	\hfill
	\begin{subfigure}[h]{0.45\textwidth}
		\centering
		\includegraphics[width=\textwidth]{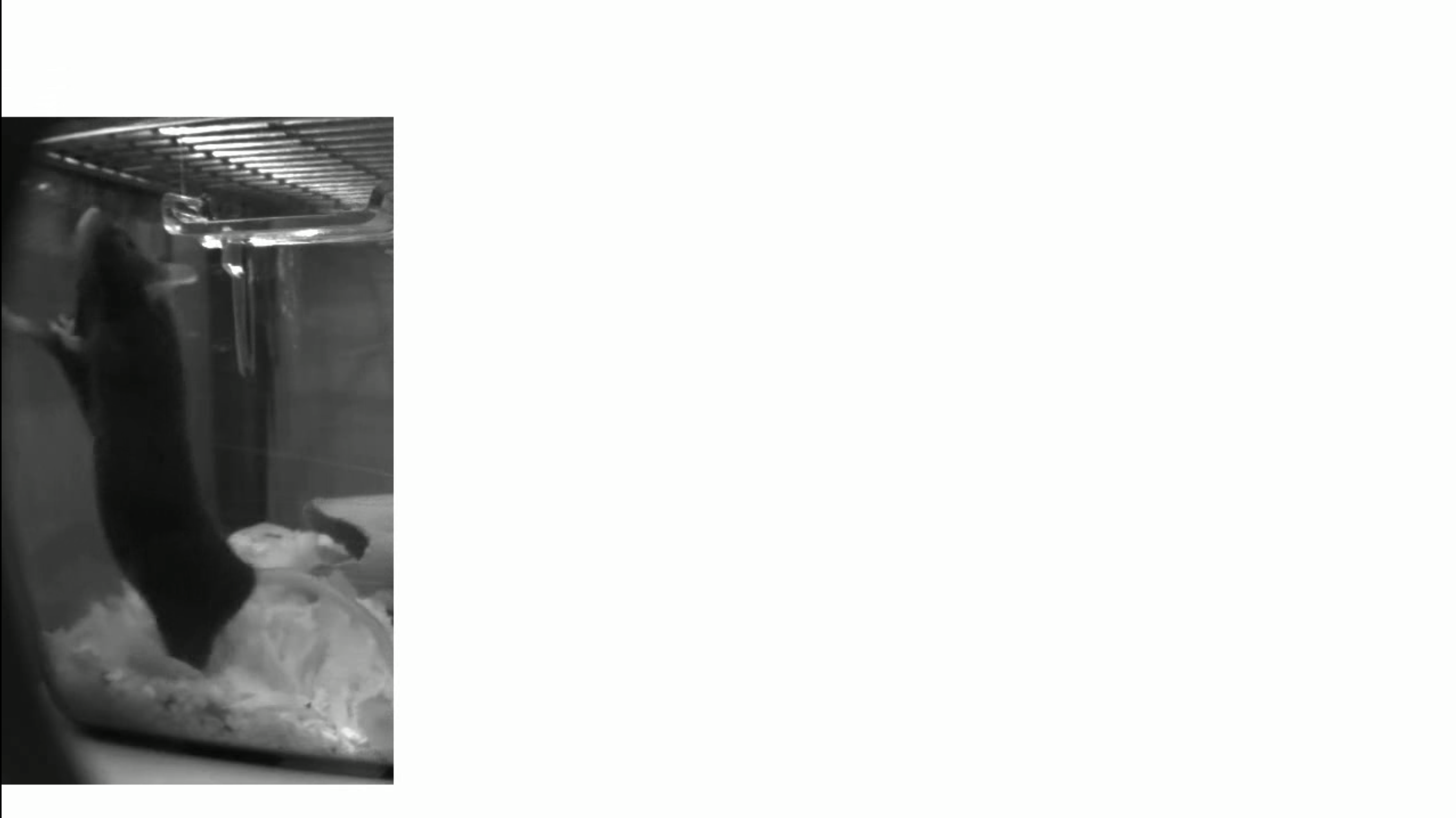}
		\caption{Example 2.}
		\label{subfig:vid_exm2}
	\end{subfigure} 
	\caption{Sample frames randomly selected from different spotlight videos.}
	\label{fig:short_vid_examples}
\end{figure}

Example frames from spotlight videos extracted using the above hyperparameter settings can be seen in figure \ref{fig:short_vid_examples}.

\begin{figure}[H]
	\centering
	\begin{subfigure}[h]{0.45\textwidth}
		\centering
		\includegraphics[width=\textwidth]{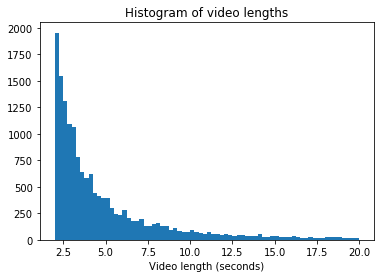}
		\caption{Range 2-> 15 seconds.}
		\label{subfig:hist2-20}
	\end{subfigure}
	\hfill
	\begin{subfigure}[h]{0.45\textwidth}
		\centering
		\includegraphics[width=\textwidth]{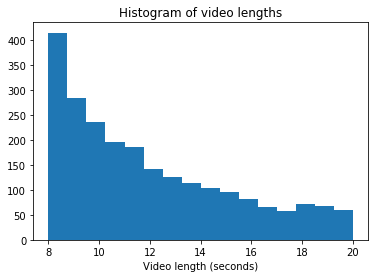}
		\caption{Range 8 -> 20 seconds.}
		\label{subfig:hist8-20}
	\end{subfigure} 
	\caption{Histogram of video lengths. Distribution contains over 16,000 videos. The distributions is truncated at 15 seconds- max video length is over 300 seconds.}
	\label{fig:histograms}
\end{figure}

\paragraph{}
Up to three videos, one tracking each rodent, can be made in parallel. However, most videos were discarded (for not having at least two seconds- 50 frames- worth of contiguous confident bounding boxes) such that each 30 minutes of video produced approximately 50-100 short (at least 2 seconds) long videos. The length of the videos, unsurpirisingly, follows a pareto distribution, see figure \ref{fig:histograms}. Here, to get a reasonably interesting time series we focus on videos of at least 8 seconds \ref{subfig:hist8-20}.

\begin{figure}[H]
    \centering
    \includegraphics[width=1.0\textwidth]{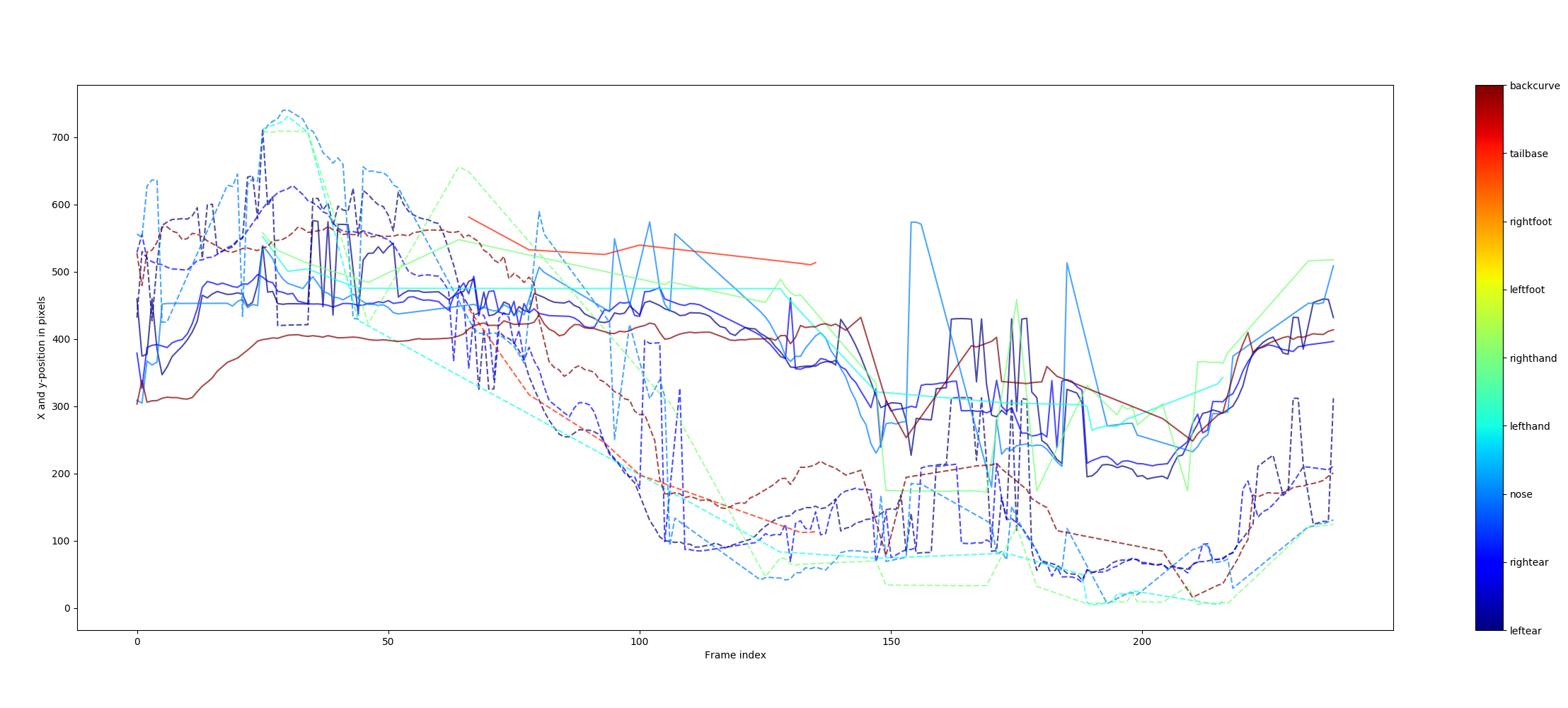}
    \caption{Trajectory analysis of labelled body parts given typical extracted video.}
    \label{fig:typical_trajectory}
\end{figure}

\paragraph{}
Now deeplabcut was trained to label the following bodyparts: leftear; rightear; snout; lefthand; righthand; leftfoot; rightfoot; tailbase; backcurve. From this labelling a position can be extracted for each bodypart in each frame. Figure \ref{fig:typical_trajectory} shows a typical trajectory analysis over a video given these positions. 

\begin{figure}[H]
    \centering
    \includegraphics[width=\textwidth]{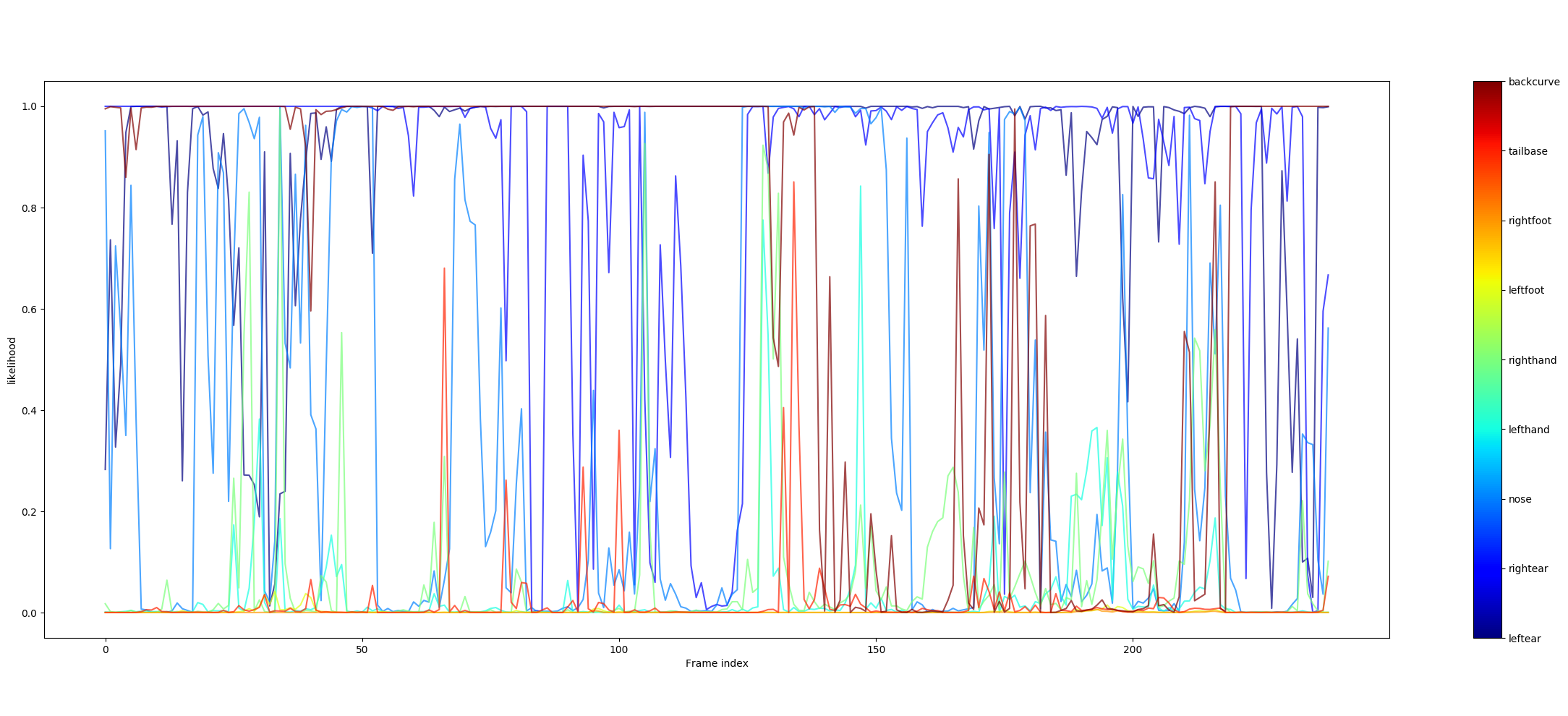}
    \caption{Likelihood analysis of labelled body parts given typical extracted video.}
    \label{fig:typical_likelihood}
\end{figure}

\paragraph{}
The positions mostly change smoothly from frame to frame, however sometimes the position suddenly moves an impossible distance which indicates an incorrect labelling likely due to partial or complete occlusion. This point is made clear by examining the likelihoods for these bodyparts of the same video as shown in figure \ref{fig:typical_likelihood}. For most videos uncertainty over bodypart positions is very erratic, thus it is impractical to determine which positions are correct and reliable from the time series and a further culling of poor videos was required. 

\subsection{Video filtering}

\paragraph{}
Creating the spotlight videos already provided a mechanism through which poor data was eliminated. However, some videos that contain a subject clearly in view may not also display the bodyparts of interest. This may occur for many reasons for example if it is hunched over facing away from the camera or most of its body is occluded by an object. For this reason we eliminate low likelihood bodyparts and design metrics to select the best videos.

\subsubsection{Threshold likelihood}

\begin{figure}[H]
	\centering
	\begin{subfigure}[h]{0.45\textwidth}
		\centering
		\includegraphics[width=\textwidth]{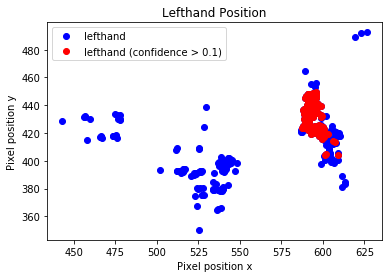}
		\caption{Position}
		\label{subfig:lefthand_position}
	\end{subfigure}
	\hfill
	\begin{subfigure}[h]{0.45\textwidth}
		\centering
		\includegraphics[width=\textwidth]{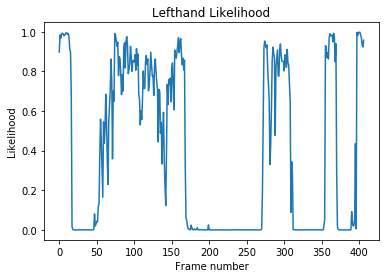}
		\caption{Likelihood}
		\label{subfig:lefthand_likelihood}
	\end{subfigure} 
	\caption{Position and likelihood of a left hand for a given video. The red points in figure \ref{subfig:lefthand_position} correspond to positions that received a greater than $0.5$ likelihood while the blue points correspond to positions that got less than $0.5$ confidence. Figure \ref{subfig:lefthand_likelihood} shows these likelihoods as a function of time.}
	\label{fig:theshold_likelihood}
\end{figure}

\paragraph{}
Deeplabcut assigns a non-zero probability to each bodypart that it has learned to detect whether the body part is visible or not in that frame as seen in figure \ref{fig:typical_likelihood}. Thus we disregard positions that have less than a certain threshold confidence. Figure \ref{subfig:lefthand_position} shows multiple clusters where all the datapoints corresponding to a higher threshold confidence are in the same cluster. This area is, by inspection, the location that contains all the correct ground truth labelling.  

\subsubsection{Threshold geomean}

\paragraph{}
The likelihood analysis (in figure \ref{fig:typical_likelihood}) suggests a good metric for selecting the highest quality videos: the high correlation between position uncertainty and low detection confidence can be mitigated by looking for videos with consistently high likelihood for the bodyparts of interest. The mean likelihood for each bodypart in each video was calculated. This metric is considered as the quality of the detection of a given bodypart for the given extracted video. Since we are interested in joint-bodypart parameterisation high quality detection of one bodypart is not useful without also obtaining high quality of the bodyparts with which its being modelled. Thus the geometric mean (equation \ref{eq:geo_mean}) of the quality of detection for different bodyparts was used as an optimisation metric for selecting the highest quality extracted videos. This more heavily penalizes outliers. Here it ensures that very high confidence in half of the bodyparts is not as large as a more equally distributed confidence.

\begin{equation}
    \mu_{geo} = \Big( \prod_{i=1}^{n} x_i \Big)^{1/n}
    \label{eq:geo_mean}
\end{equation}

\begin{figure}[H]
    \centering
    \includegraphics[width=\textwidth]{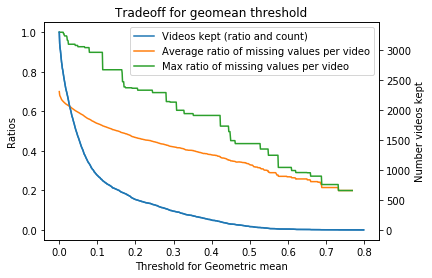}
    \caption{Tradeoff between geomean threshold and fraction of videos kept and the fraction of total missing datapoints in the remaining videos.}
    \label{fig:threshold_geomean}
\end{figure}

Figure \ref{fig:threshold_geomean} shows the ratio of videos kept, the average number of datapoints missing (which mean they had a confidence of less than 0.5) in the kept videos and the maximum fraction of datapoints missing out of the videos kept as a function of the geometric mean threshold of acceptance.

\paragraph{}
Here we selected a geomean threshold of $0.3$ for the bodyparts: leftear, rightear, snout, lefthand, and righthand. The resulting videos and their coordinates for these bodyparts were then progressed to the differential smoothing and interpolation stage discussed below in section \ref{sec:interpolation}.

\section{Differential Smoothing and Interpolation}
\label{sec:interpolation}

\begin{figure}[H]
    \centering
    \includegraphics[width=\textwidth]{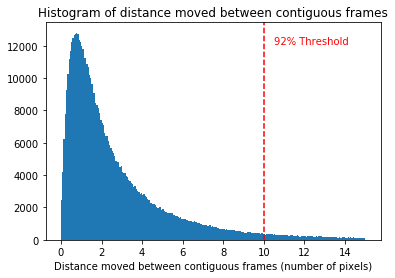}
    \caption{Histogram of the distance in pixels moved between contiguous frames for all bodyparts in selected videos. The maximum allowed distance was set to $\sqrt{x^2 + y^2} = 10.0$ perserving 92\% of the datapoints. This is eliminating the most spurious datapoints. Non-spurious datapoints eliminated here are interpolated very well.}
    \label{fig:diffs_threshold}
\end{figure}

\paragraph{}
Now that we have a method of selecting the videos with least occlusions and highest confidences for the joint bodyparts of interest we must construct a sensible time series from them. While interpolating it would be optimal to, as much as possible, eliminate single datapoints where the position of the associated bodypart moves by a large amount. We do this by considering the euclidean distance ($\sqrt{x^2 + y^2}$) between the pixel positions of bodyparts in contiguous frames. This distribution for all videos and all bodyparts is shown in figure \ref{fig:diffs_threshold}. We set the maximum allowed movement to be $\sqrt{x^2 + y^2} = 10.0$. If the change was greater than this the position in the latter frame was set to uncertain (nan). This threshold keeps 92\% of the datapoints.

\begin{figure}[H]
	\centering
	\begin{subfigure}[h]{0.45\textwidth}
		\centering
		\includegraphics[width=\textwidth]{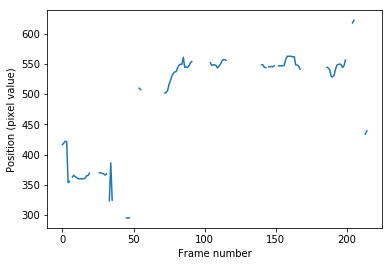}
		\caption{Non-smoothed}
		\label{subfig:time_series_original}
	\end{subfigure}
	\hfill
	\begin{subfigure}[h]{0.45\textwidth}
		\centering
		\includegraphics[width=\textwidth]{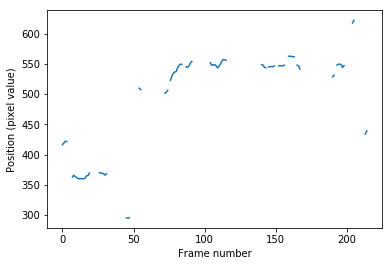}
		\caption{With differential smoothing.}
		\label{subfig:time_series_smoothed}
	\end{subfigure} 
	\caption{Time series of an example bodypart for an example video. Smoothing eliminates most sharp spikes. Particularly noticeable at about frame 40.}
	\label{fig:diffs}
\end{figure}

\paragraph{}
An example of a non-smoothed and smoothed time series can be seen in figure \ref{subfig:time_series_original} and \ref{subfig:time_series_smoothed} respectably. Since we shall interpolate over the missing points this will mostly only affect spurious datapoints and genuine rapid changes shall be respected by interpolation.

\begin{figure}[H]
    \centering
    \includegraphics[width=\textwidth]{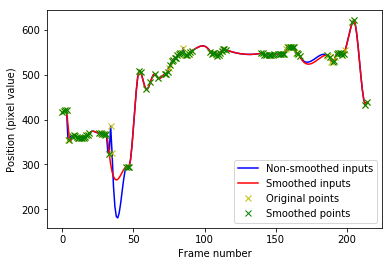}
    \caption{Interpolation with a cubic spline overlayed with and without the smoothing pre-processing. In regions where large changes were non-spurious the two interpolations are almost identical. However in regions where large changes were spurious a significantly more reasonable interpolation was achieve.}
    \label{fig:interpolated_time_series}
\end{figure}

\paragraph{}
The smoothing process clearly eliminates spurious sharp changes in position. It also deletes multiple reasonable leaps (see the region between frame 150 and 170). However, in regions where these leaps were reasonable interpolating with a cubic spline gives an almost identical time series. To illustrate this point we overlay two interpolations, one without smoothing and one with smoothing. This can be seen in figure \ref{fig:interpolated_time_series}.

\paragraph{}
We can see that the region between frame 150 to 170 is interpolated in almost an identical way- no effective information has been lost. However, the interpolation in the region from 40-50 is much more reasonable for the inputs which have been differentially smoothed due to the removal of one outlying spurious datapoint in this region which forces a much more unnatural curve.

\paragraph{}
Finally, for the interpolated time series the bodypart positions for each frame of the spotlight videos are normalised by the value of of the centroid of the bodyparts for that frame. This makes captured behaviour spatially independent.

\newpage

\section{Manifold Approximation and Projection}

\subsection{Motivation of Uniform Manifold Approximation and Projection}

UMAP (Uniform Manifold Approximation and Projection) is a relatively new manifold learning technique for dimension reduction which is constructed from a theoretical framework based in Riemannian geometry and algebraic topology \cite{umap}. UMAP is particularly relevant here since, unlike t-SNE, UMAP has no computational restrictions on embedding dimension, making it viable as a general purpose and non-linear dimension reduction technique.

\subsection{Behavioural Windows from Spotlight videos}

\paragraph{}
Once the interpolation outlined in section \ref{sec:interpolation} was completed the result is many multiple dimensional time series, one for each spotlight video selected in section \ref{sec:spotlight}. We now extract multi-dimensional time series windows similar to \cite{timecluster}, where each window has dimension $\omega \times 2f$ where $f$ is the number of bodyparts of interest each of which is a 2-dimensional cartesian coordinate and $\omega=60$ is the the window size in number of frames. We also specify a window stride $s=1$ such that for a video with 100 frames we have 40 windows: $[1,60], [2,61], \cdots [61,100]$. Each of these windows is now a high dimensional representation of a behaviour. We wish to now express the relationship between these behaviours visually. 

\subsection{2D Projections and inspecting clusters}

\begin{figure}[H]
    \centering
    \includegraphics[width=\textwidth]{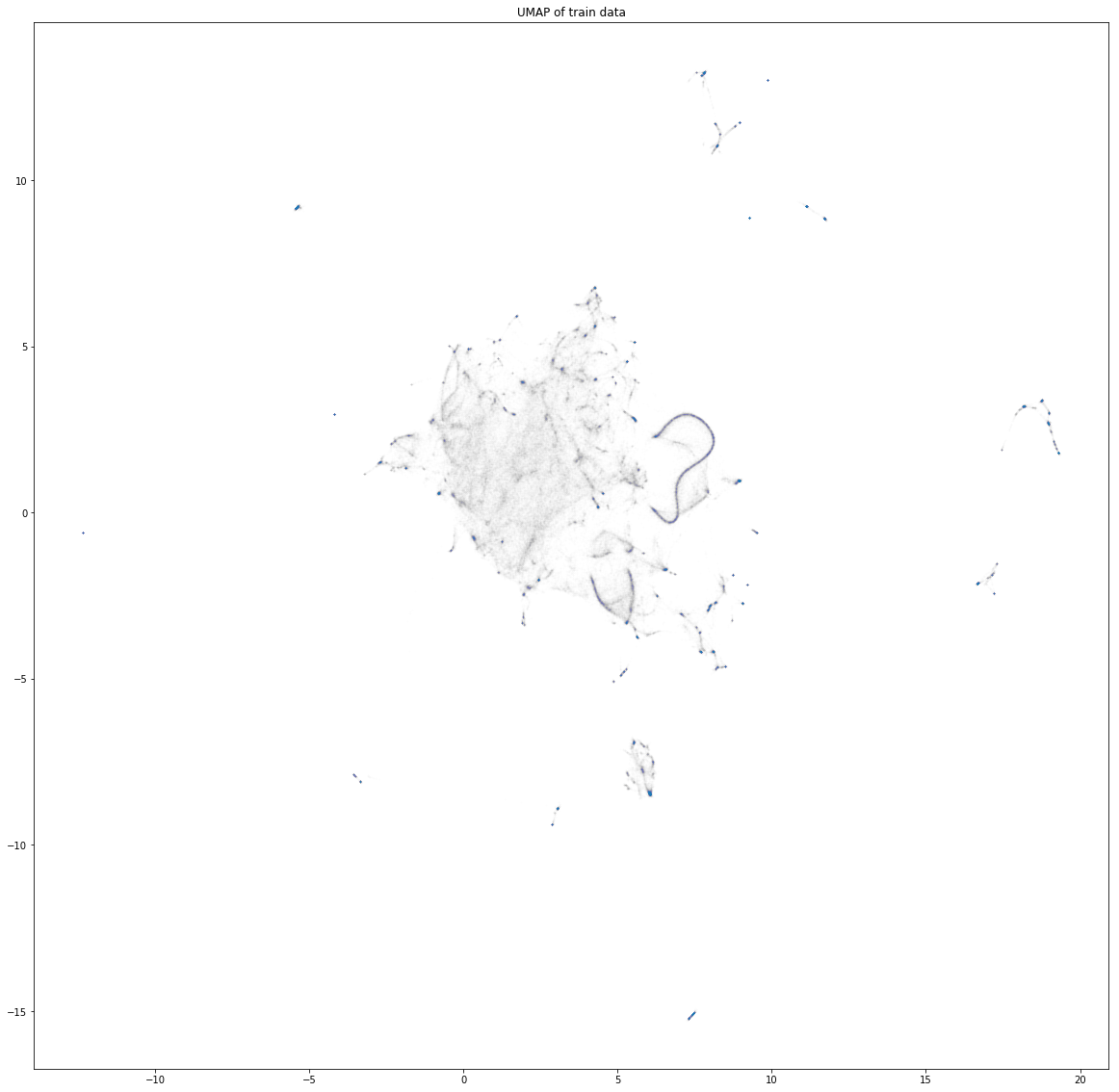}
    \caption{2 dimensional projection of behavioural windows for over 300 selected spotlight videos using UMAP with number of neighbours = 200 and minimum distance between points = 0.0. $\omega=60$, $s=1$. The points' size and transparency is minimised thus highly dense clusters represent very common stereotyped behaviour. More faint points connecting dense clusters are likely state transitions, very brief and not necessarily precisely replicable. These stereotyped clustered have been produced in an entirely unsupervised way. However, a labelling of clusters is easily obtained by inspection.}
    \label{fig:umap_train}
\end{figure}

\paragraph{}
A train set of 200,000 behavioural windows was used to train UMAP and a further 20,000 were used for cross validation to select optimal hyperparameters for UMAP. This search was over number of neighbours (in the region 1 to 200) and minimum distance between points (in the region $0.0$ to $1.0$). The optimal mapping, determined by the fromation of the most distinct clusters, is shown in figure \ref{fig:umap_train}. Each point now corresponds to a behaviourial window. We can see dense clustering representing stereotyped behaviour where similar behaviour tends to be in the same cluster. More irregular, or fast, behaviour such as state transitions can be seen connecting these dense clusters. 

\paragraph{}
This mapping was compared to a baseline linear dimensionality reduction mechanism, Primary Component Analysis (PCA). The only clearly distinct clusters formed in this way differentiated between behaviour where there was little to no movement, such as sleep, from most other waking behaviour. This hence demonstrated the foundational non-linear relationship between rodent pose and stereotyped behaviour. Preliminary dimensionality reduction simulations also included the used of a Deep Convolutional Auto-Encoder (DCAE) to create a latent space of 60 dimensions and then using UMAP to reduce further to 2 dimensions, similarly to \cite{timecluster}. This gave a significant improvement over the PCA baseline however did not result in clusters as distinct as applying UMAP directly on the the time series windows.  

\begin{figure}[H]
    \centering
    \includegraphics[width=\textwidth]{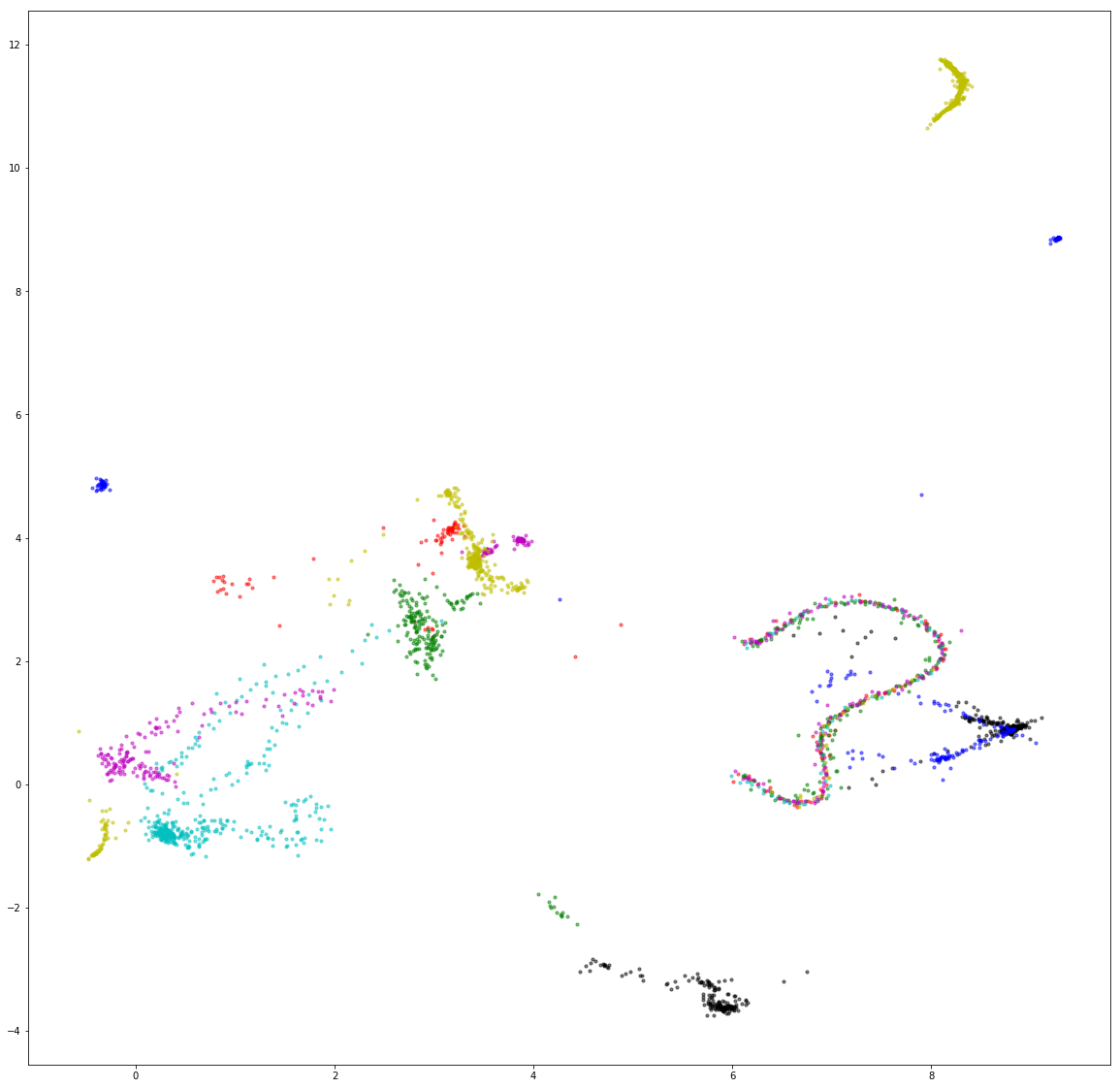}
    \caption{8 videos all identified, by inspection, to only contain the behaviour of reaching out for for food and then eating it from their hands. Distinctly different amongst these behaviours is blue and black that correspond to videos where we have this same scenario except the subject is standing on top of cardboard fun tunnel while doing this causes it to reach out for food in a distinctly different way as can be seen from the distinct arrow like shape to the right of the ear-like cluster.}
    \label{fig:eating_multiple}
\end{figure}

\paragraph{}
The ear-like structure toward the right centre, by inspection, corresponds to a rodent stretched out grabbing food. Different segments of this ear structure are slight different movements of the hands and snout while doing this. Note that while the window size is $\omega=60$ corresponding to just over 2 seconds, the time scale of behaviour in reality is much more sensitive than this. A stride of $s=1$ ensures this overlap and captures the transition from one state to the next. Examining videos corresponding to the ear-like cluster more closely we see that videos where the subject is eating for the whole video follows a distinct pattern. This at a course grain (human-like) level of stereotyping has two distinct states. See figure \ref{fig:eating_multiple}.

\paragraph{}
All colours (each corresponding to one of 8 distinct videos labelled retrospectively as eating) except blue and black have points in the ear like structure as well as a second cluster somewhere to the left of this. These two course grain locations correspond to the human-identifiable positions of reaching for food and eating the snatched food from hands. The subtle difference in pose of this reach when doing so while standing on top of cardboard fun tunnel, the blue and the black, are also captured. However, this behavioural analysis gives a broader variety of analysis in way in which food is eaten from hands as can be seen from the distinctly different clusters to the left. 

\begin{figure}[H]
    \centering
    \includegraphics[width=\textwidth]{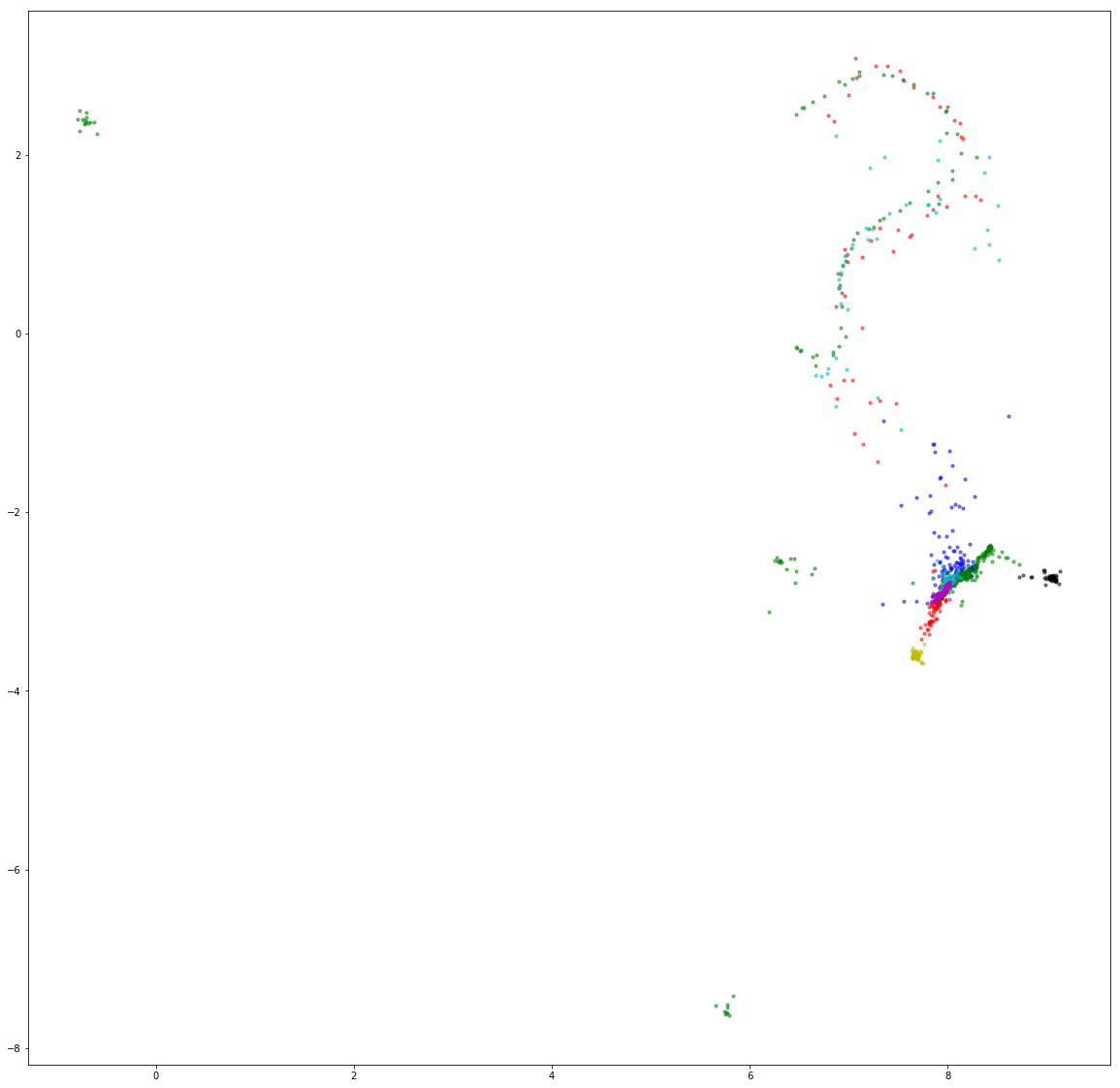}
    \caption{6 videos all identified, by inspection, to only contain the behaviour of licking the water bottle. Distinctly different amongst these behaviours is green in which the subject climbs over cardboard fun tunnel just before drinking. In some of these videos the subject occasionally briefly reaches to the food above the water bottle highlighting a few points in the ear-like cluster.}
    \label{fig:drinking_multiple}
\end{figure}

\paragraph{}
The drinking cluster is very densely packed. To the human eye the poses of a subject licking the water bottle is almost uni-modal. However here the subtle change in posture is captured by the spatial distinction in these clusters, black being the most dissimilar posture from red or yellow.

\begin{figure}[H]
	\centering
	\begin{subfigure}[h]{0.45\textwidth}
		\centering
		\includegraphics[width=\textwidth]{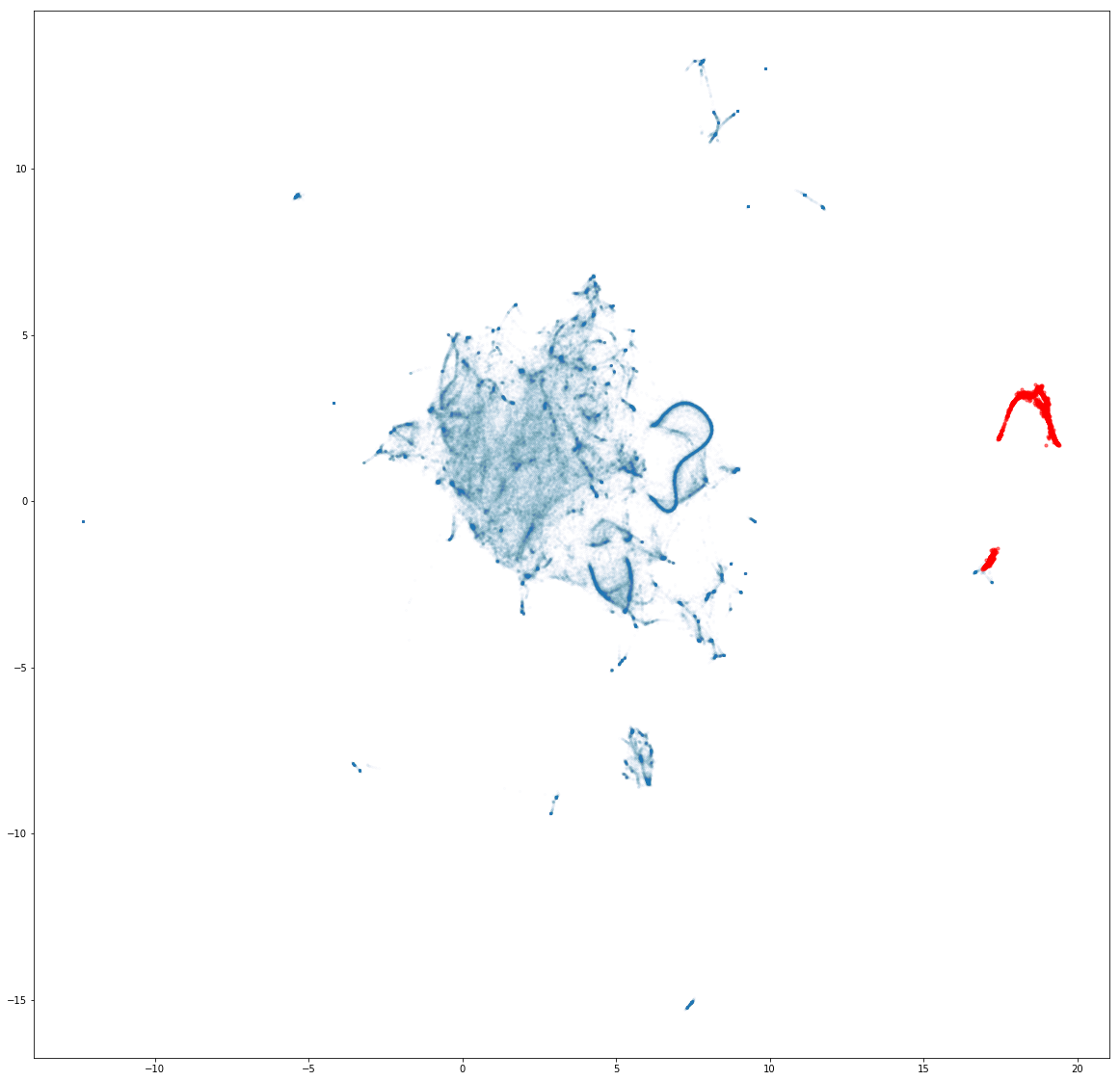}
		\caption{Grooming.}
		\label{subfig:grooming}
	\end{subfigure}
	\hfill
	\begin{subfigure}[h]{0.45\textwidth}
		\centering
		\includegraphics[width=\textwidth]{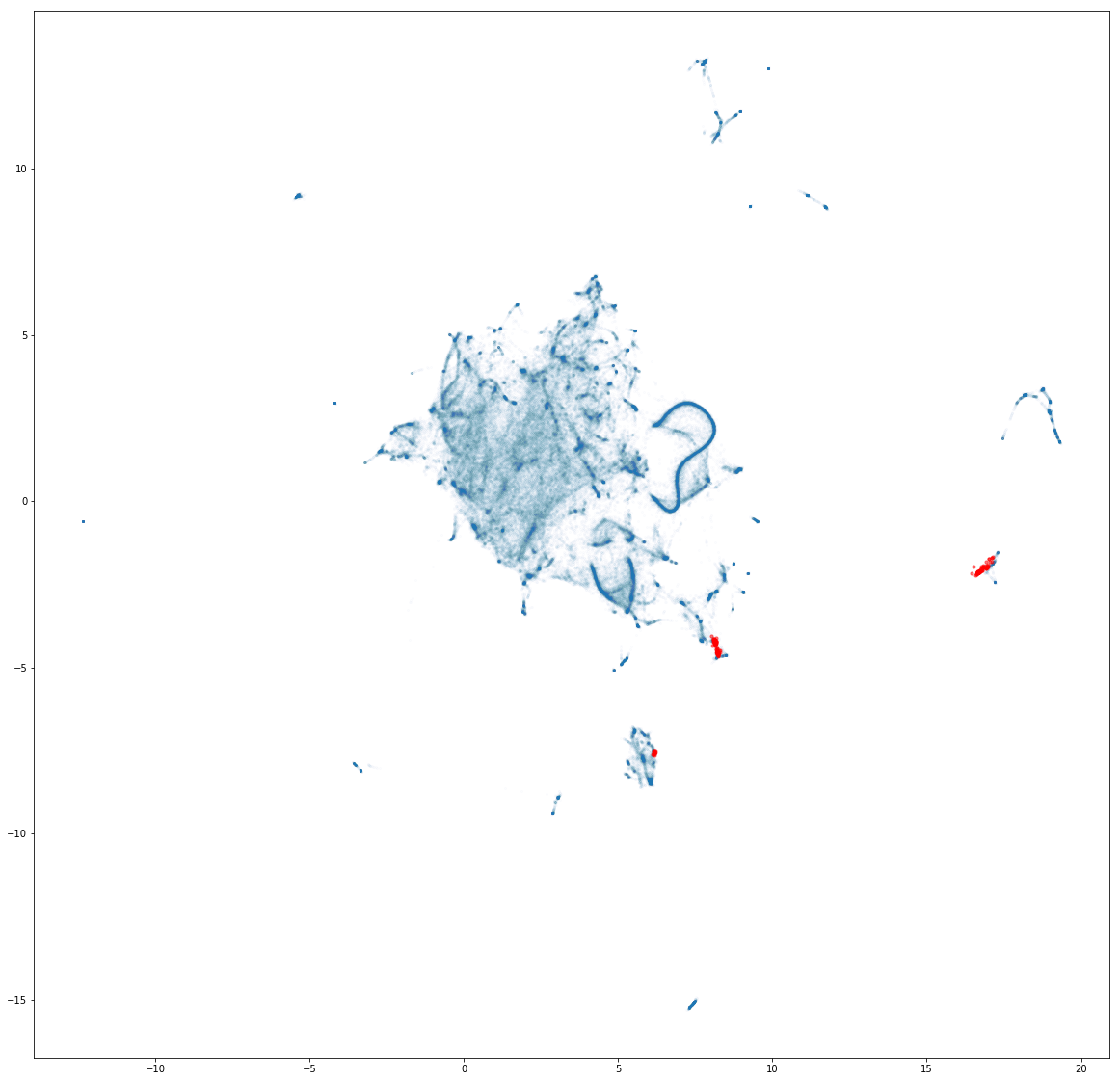}
		\caption{Grooming.}
		\label{subfig:grooming2}
	\end{subfigure} 
	\begin{subfigure}[h]{0.45\textwidth}
		\centering
		\includegraphics[width=\textwidth]{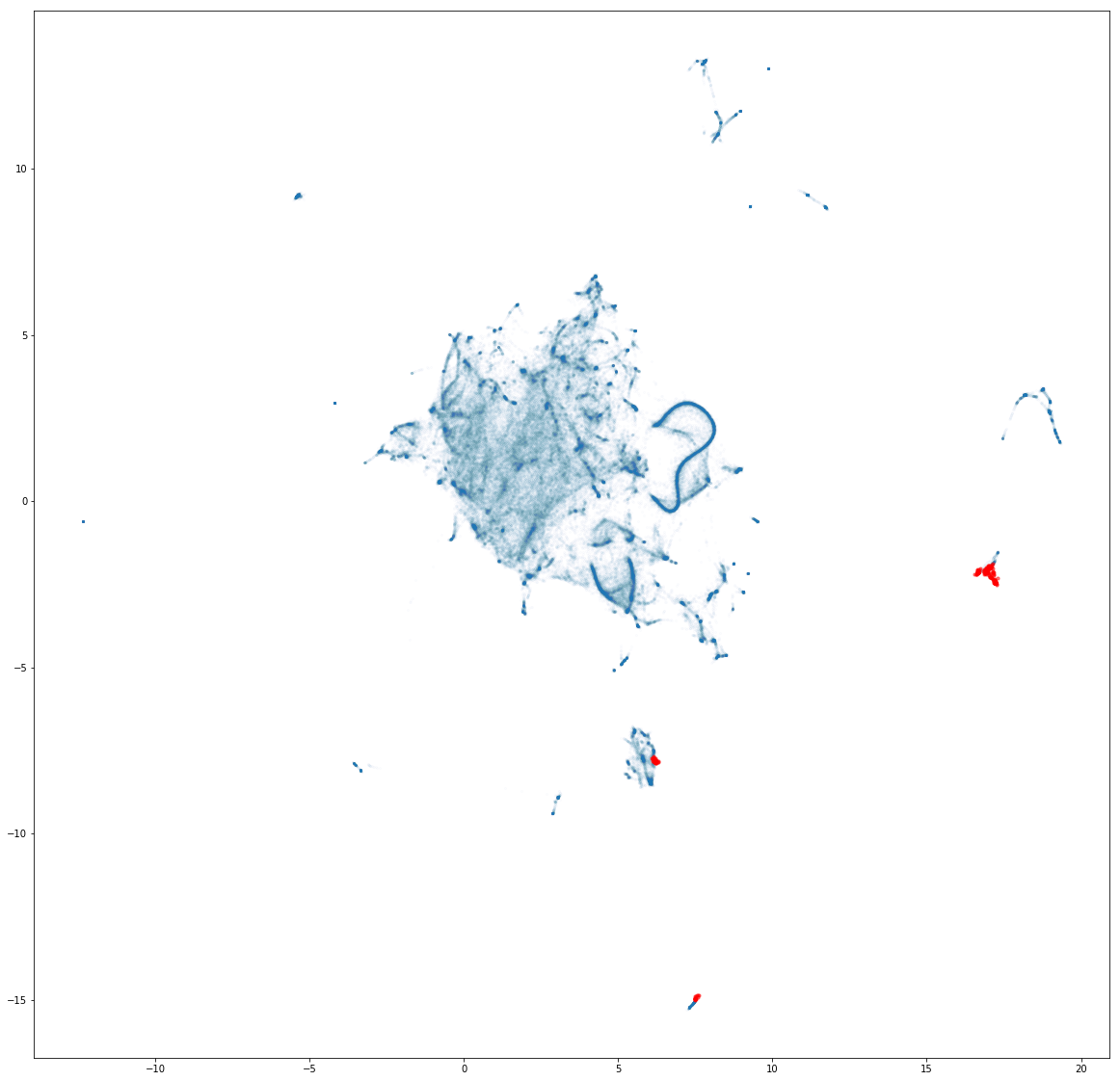}
		\caption{Grooming.}
		\label{subfig:grooming3}
	\end{subfigure}
	\hfill
	\begin{subfigure}[h]{0.45\textwidth}
		\centering
		\includegraphics[width=\textwidth]{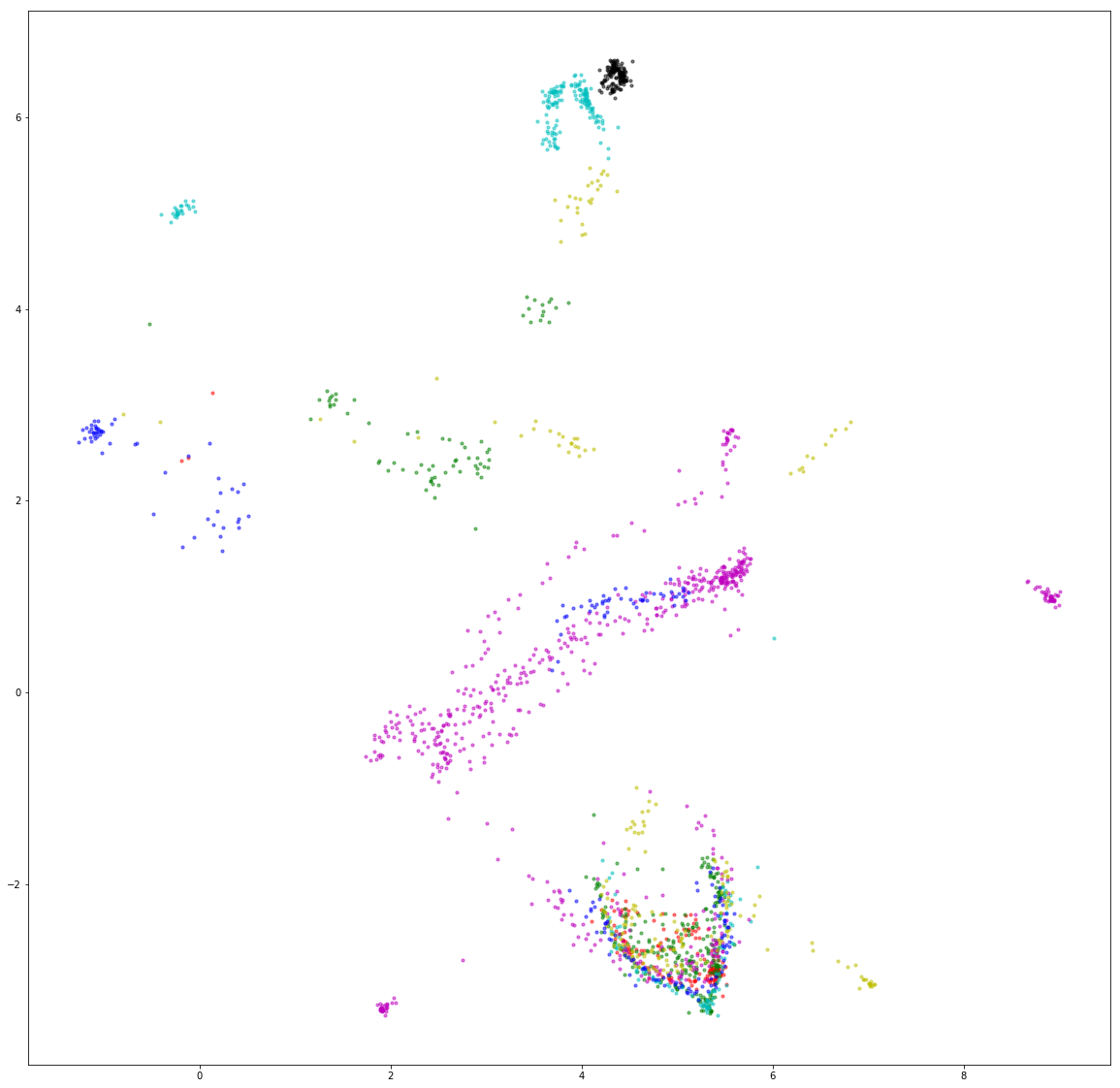}
		\caption{Eating while corached, usually because there's a second rodent competing for space.}
		\label{subfig:eating_croached}
	\end{subfigure} 	
	\caption{4 more clusters labelled by inspection. Figure \ref{subfig:eating_croached} can be compared to figure \ref{fig:eating_multiple}. It is also videos of a rodent reaching out for food but from a distinctly more crouched position usually because there is another rodent behind it competing for space. Here, similarly to figure \ref{fig:eating_multiple} there are two distinct human identifiable stages to the behaviour. One where the subject reaches out for food- the horse shoe like cluster- and a second where it eats from its hands in multiple subtly different ways.}
	\label{fig:clusters_multiple}
\end{figure}

\paragraph{}
Multiple distinct human-identifiable behaviours can be seen on other dense clusters such as grooming, burrowing and sleeping. Some examples can be seen in figure \ref{fig:clusters_multiple}. However, the difference in behaviour for most dense clusters are very subtle and some may contain distinct systematic bias. In figure \ref{fig:umap_train} we can see faint grey trails between many of the point like dense clusters seem to represent very brief state transitions between the stereotyped dense clusters.

\subsection{Edge Detection and Ensemble}

\begin{figure}[H]
    \centering
    \includegraphics[width=\textwidth]{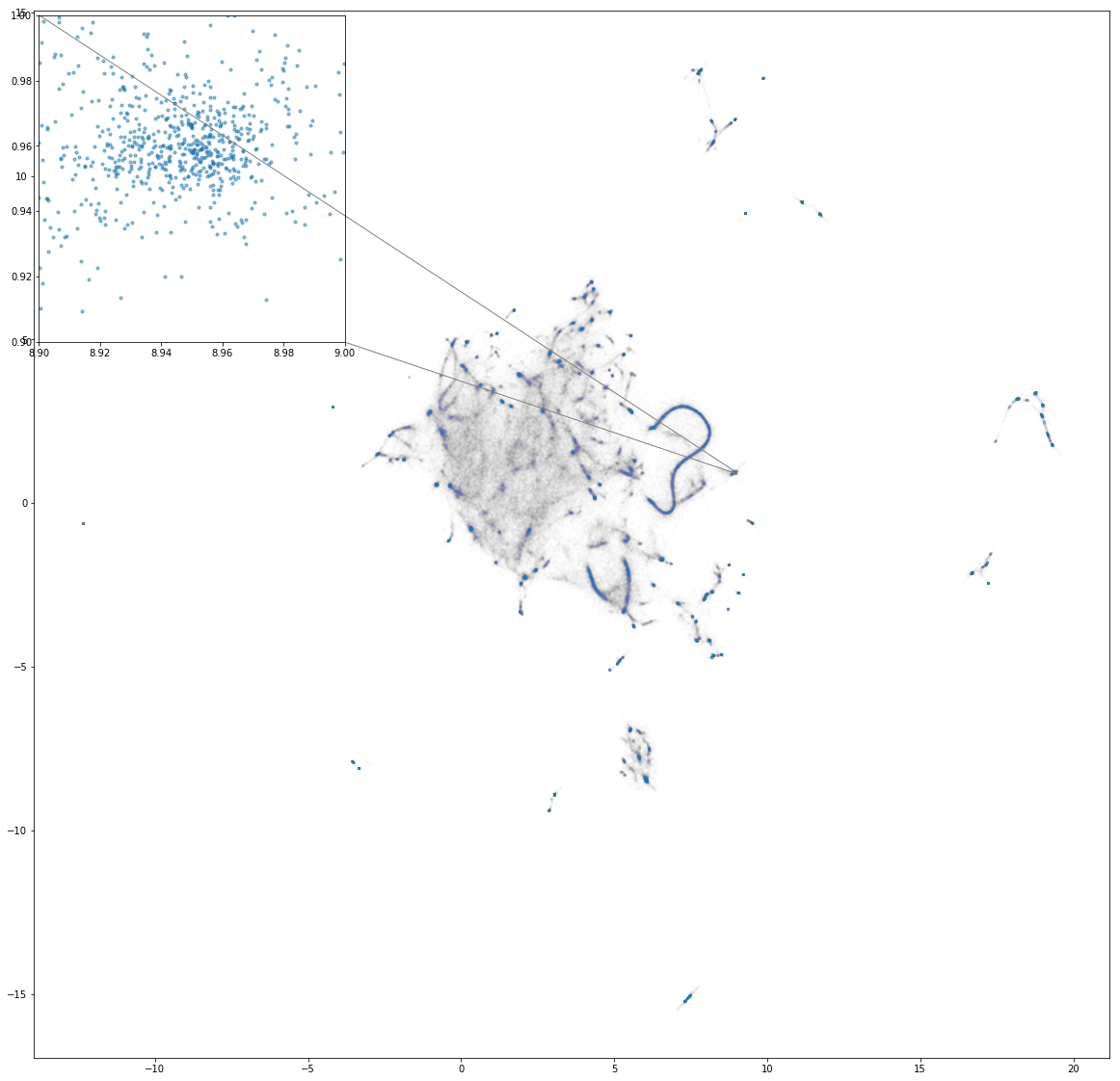}
    \caption{Dense behaviour cluster selected to examine. By inspection this cluster corresponds to the subject eating while standing on a cylindrical obstacle (cardboard fun tunnel).}
    \label{fig:eating_on_bog_roll}
\end{figure}

\paragraph{}
To demonstrate the stereotyped behaviour belonging to cluster locations we performed canny edge detection \cite{canny} on each frame in each behavioural window corresponding to a particular spatial location, see figure \ref{fig:eating_on_bog_roll}. 

\paragraph{}
The pixel values in the resulting edge detected frames were averaged across all corresponding frames in other behaviour windows within the selected spatial location. This resulted in videos of of stereotyped behaviour where the outline of the subject, along with the static background, can be clearly seen.

\begin{figure}[H]
    \centering
    \includegraphics[width=\textwidth]{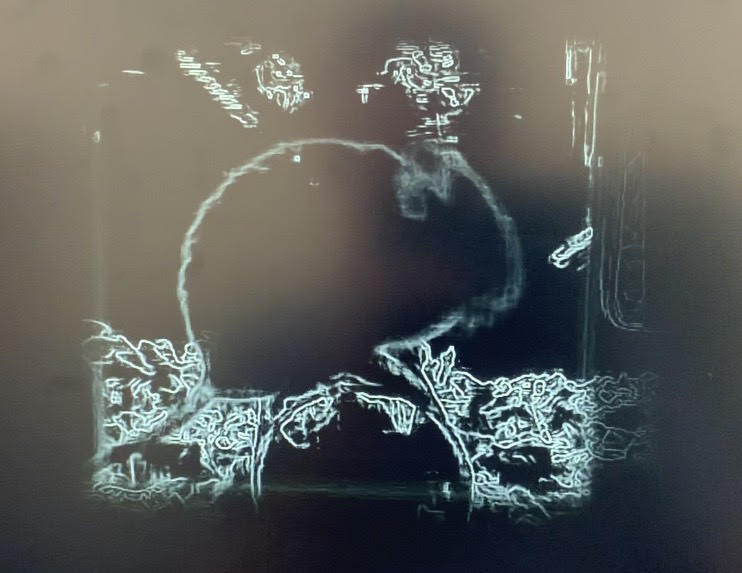}
    \caption{A sample frame from the Edge Detection and Ensemble videos. After applying canny edge detection, the pixel values have been averaged and normalised across 367 corresponding frames from behavioural windows within the same spatial cluster (figure \ref{fig:eating_on_bog_roll}) thus only the precise pose that corresponds to this section of the behavioural window within the spatial cluster can be seen.}
    \label{fig:edge}
\end{figure}

\paragraph{}
Figure \ref{fig:edge} shows a single frame from a stereotyped video made using this edge detection and ensemble technique of the spatial location highlighted in figure \ref{fig:eating_on_bog_roll}. Since the pixel values of the edge detection was averaged and normalised over 367 videos only pixel locations that consistently represented the edge of an object remain. Choosing too large spatial locations makes the outline of the subject less localised and more faint. The area needs to be very tight to get a clear outline thus demonstrating the precision of the 2 dimensional embedding.

\paragraph{}
Juxtaposing the sharpness of the edges of the subject's body with the edges of the static background a slightly blur can be identified. This represents the small variation in position since although we have examined a very small location in the embedding space it is not infinitesimal
so the subject's position will vary slightly while the static background should not.

\section{Conclusions and Future improvements}

\paragraph{}
These results provide excellent proof of concept of the potential of markerless unsupervised visual analysis. Resulting visualisations can clearly be used for anomaly detection. Comparison of non-linear UMAP to a linear PCA demonstrated the inherent complex non-linear relationship between pose and behaviour. A further extension would be to train a Variational Dynamical Encoder (VDE), as in \cite{vde}. Here a latent space representation of a behavioural window can be used to predict a latent space of a behavioural window at some stride, $s$, later. Anomalies can then be detected when the probability of one state transferring into another is sufficiently low.

\paragraph{}
Ali et al. \cite{timecluster} form a simple 2-dimensional representation of uni and multidimensional time series data. For multi-dimensional time series they used a Deep Convolutional Auto-Encoder (DCAE) to learn a 60 dimensional representation without supervision. They then reduced the dimensionality further using PCA, t-SNE and UMAP to obtain a two dimensional representation and the subsequent visualisation. As discussed, in our preliminary analysis we used a similar approach to \cite{timecluster}, using a DCAE to obtain a 60 dimensional latent space and then UMAP to reduce to a visualisable 2 dimensions. However, UMAP alone proved a powerful enough tool to be able to justify not using DCAE as an initial reduction and in fact resulted in more distinct behavioural clusters. However, there is a very large number of architectures and hyperparameters that can be tried here to achieve better results.

\paragraph{}
Here we considered a behavioural window size of $\omega=60$ which is just over two seconds. Although a stride size of $s=1$ makes it still possible to capture changes at a much finer timescale, behaviour over much greater time periods may be captured by varying this value. 

\paragraph{}
The restriction of low data quality and limited by lack of pose estimation systems for rodents emphasised the importance of accurate pose extraction for the purpose of complex unsupervised videographic behavioural analysis. We were able to demonstrate the 
existence of stereotyped behaviour visually by creating comprehensive embedding space. However, this required a very complex set of pre-processing steps to extract data of high enough quality such as not to have spurious time series. Thus, such a system will be poor at inference and may only be improved by more sophisticated pose estimation for rodents. As discussed in the introduction, such more advanced systems have already been developed for humans only \cite{openpose}, \cite{densepose}. Thus, unsupervised videographic analysis of human behaviour is easier to approach from the perspective of comprehensive pose extraction support. Moreover, since humans have a much more complex myriad of behavioural stereotypes, particularly at different time scales, the potential for far deeper behavioural insights as well as ultra impactful applications such as healthcare, crime, and commerce are tremendous.

\bibliographystyle{unsrt}  
\bibliography{main}  

%
%
%
%

\end{document}